\documentclass[final]{cvpr}

\usepackage{times}
\usepackage{epsfig}
\usepackage{graphicx}
\usepackage{subfigure}

\usepackage{amsmath,amssymb,amsfonts,dsfont,bm,mathrsfs,pifont}
\usepackage{booktabs,multirow,adjustbox}
\usepackage[pagebackref=true,breaklinks=true,colorlinks,bookmarks=false]{hyperref}
\newcommand{\z}{{\rm\bf z}}      
\newcommand{\Z}{\mathcal{Z}}     
\newcommand{\w}{{\rm\bf w}}      
\newcommand{\W}{\mathcal{W}}     
\newcommand{\y}{{\rm\bf y}}      
\newcommand{\Y}{\mathcal{Y}}     
\newcommand{\x}{{\rm\bf x}}      
\newcommand{\f}{{\rm\bf f}}      
\newcommand{\E}{\mathbb{E}}      
\newcommand{\FID}{FID$\downarrow$}  
\newcommand{\SSIM}{SSIM$\uparrow$}  
\newcommand{\MSE}{MSE$\downarrow$}  

\newlength\savewidth

\newcommand{\blocks}[3]{\multirow{3}{*}{\(\left[\begin{array}{c}\text{1$\times$1, #2}\\[-.1em] \text{3$\times$3, #2}\\[-.1em] \text{1$\times$1, #1}\end{array}\right]\)$\times$#3}
}

\begin{document}

\title{Generative Hierarchical Features from Synthesizing Images}

\author{
  Yinghao Xu\thanks{denotes equal contribution.} \quad  Yujun Shen\footnotemark[1] \quad Jiapeng Zhu \quad  Ceyuan Yang \quad Bolei Zhou \\
  The Chinese University of Hong Kong \\
  {\tt\small \{xy119, sy116, jpzhu, yc019, bzhou\}@ie.cuhk.edu.hk}
}

\maketitle

\begin{abstract}
Generative Adversarial Networks (GANs) have recently advanced image synthesis by learning the underlying distribution of the observed data.
However, how the features learned from solving the task of image generation are applicable to other vision tasks remains seldom explored.
In this work, we show that learning to synthesize images can bring remarkable hierarchical visual features that are generalizable across a wide range of applications.
Specifically, we consider the pre-trained StyleGAN generator as a learned loss function and utilize its layer-wise representation to train a novel hierarchical encoder.
The visual feature produced by our encoder, termed as Generative Hierarchical Feature (GH-Feat), has strong transferability to both generative and discriminative tasks, including image editing, image harmonization, image classification, face verification, landmark detection, and layout prediction.
Extensive qualitative and quantitative experimental results demonstrate the appealing performance of GH-Feat.%
\footnote{Project page is at \url{https://genforce.github.io/ghfeat/}.}
\end{abstract}

\section{Introduction}\label{sec:introcution}

Representation learning plays an essential role in the rise of deep learning.
The learned representation is able to express the variation factors of the complex visual world.
Accordingly, the performance of a deep learning algorithm highly depends on the features extracted from the input data.
As pointed out by Bengio \textit{et al.}~\cite{bengio2013representation}, a good representation is expected to have the following properties.
First, it should be able to capture multiple configurations from the input.
Second, it should organize the explanatory factors of the input data as a hierarchy, where more abstract concepts are at a higher level.
Third, it should have strong transferability, not only from datasets to datasets but also from tasks to tasks.

Deep neural networks supervisedly trained for image classification on large-scale datasets (\textit{e.g.}, ImageNet~\cite{imagenet} and Places~\cite{zhou2017places}) have resulted in expressive and discriminative visual features~\cite{sharif2014cnn}.
However, the developed features are heavily dependent on the training objective.
For example, prior work has shown that deep features trained for the object recognition task may mainly focus on the shapes and parts of the objects while remain invariant to rotation~\cite{bau2017network,matthew2014visualizing}, and the deep features from a scene classification model may focus more on detecting the categorical objects (\textit{e.g.}, bed for bedroom and sofa for living room)~\cite{zhou2015object}.
Thus the discriminative features learned from solving high-level image classification tasks might not be necessarily good for other mid-level and low-level tasks, limiting their transferability~\cite{yosinski2014transferable,zhao2020makes}.
Besides, it remains unknown how the discriminative features can be used in generative applications like image editing.

Generative Adversarial Network (GAN)~\cite{gan} has recently made great process in synthesizing photo-realistic images.
It considers the image generation task as the training supervision to learn the underlying distribution of real data.
Through adversarial training, the generator can capture the multi-level variations underlying the input data to the most extent, otherwise, the discrepancy between the real and synthesized data would be spotted by the discriminator.
The recent state-of-the-art StyleGAN~\cite{stylegan} has been shown to encode rich hierarchical semantics in its layer-wise representations~\cite{stylegan,higan,interfacegan}.
However, the generator is primarily designed for image generation and hence lacks the inference ability of taking an image as the input and extracting its visual feature, which greatly limits the applications of GANs to real images.

To solve this problem, a common practice is to introduce an additional encoder into the two-player game described in GANs~\cite{bigan,ali,bigbigan,alae}.
Nevertheless, existing encoders typically choose the initial latent space (\textit{i.e.}, the most abstract level feature) as the target representation space, omitting the pre-layer information learned by the generator.
On the other hand, the transferability of the representation from GAN models is not fully verified in the literature.
Most prior work focuses on learning discriminative features for the high-level image classification task~\cite{bigan,ali,bigbigan} yet put little effort on other mid-level and low-level downstream tasks, such as landmark detection and layout prediction.

In this work, we show that the pre-trained GAN generator can be considered as a learned loss function.
Training with it can bring highly competitive hierarchical visual features which are generalizable to various tasks.
Based on the StyleGAN model, we tailor a novel hierarchical encoder whose outputs align with the layer-wise representations from the generator.
In particular, the generator takes the feature hierarchy produced by the encoder as the per-layer inputs and supervises the encoder via reconstructing the input image.
We evaluate such visual features, termed as \textit{Generative Hierarchical Features (GH-Feat)}, on both generative and discriminative tasks, including image editing, image harmonization, image classification, face verification, landmark detection, layout prediction, \textit{etc}.
Extensive experiments validate that the generative feature learned from solving the image synthesis task has compelling hierarchical and transferable properties, facilitating many downstream applications.

\section{Related Work}\label{sec:related-work}

\noindent\textbf{Visual Features.}
Visual Feature plays a fundamental role in the computer vision field.
Traditional methods used manually designed features~\cite{sift,surf,hog} for pattern matching and object detection.
These features are significantly improved by deep models~\cite{alexnet,vgg,resnet}, which automatically learn the feature extraction from large-scale datasets.
However, the features supervisedly learned for a particular task could be biased to the training task and hence become difficult to transfer to other tasks, especially when the target task is too far away from the base task~\cite{yosinski2014transferable,zhao2020makes}.
Unsupervised representation learning is widely explored to learn a more general and transferable feature~\cite{doersch2017multi,zhang2017split,wu2018unsupervised,gidaris2018unsupervised,hjelm2019learning,zhuang2019local,moco,cpc,cpc2,cmc}.
However, most of existing unsupervised feature learning methods focus on evaluating their features on the tasks of image recognition, yet seldom evaluate them on other mid-level or low-level tasks, let alone generative tasks.
Shocher \textit{et al.}~\cite{shocher2020semantic} discover the potential of discriminative features in image generation, but the transferability of these features are still not fully verified.

\vspace{5pt}
\noindent\textbf{Generative Adversarial Networks.}
GANs~\cite{gan} are able to produce photo-realistic images via learning the underlying data distribution.
The recent advance of GANs~\cite{dcgan,pggan,biggan} has significantly improved the synthesis quality.
StyleGAN~\cite{stylegan} proposes a style-based generator with multi-level style codes and achieves the start-of-the-art generation performance.
However, little work explores the representation learned by GANs as well as how to apply such representation for other applications.
Some recent work interprets the semantics encoded in the internal representation of GANs and applies them for image editing~\cite{gansteerability,interfacegan,gandissect,mganprior,higan,idinvert}.
But it remains much less explored whether the learned GAN representations are transferable to discriminative tasks.

\vspace{5pt}
\noindent\textbf{Adversarial Representation Learning.}
The main reason of hindering GANs from being applied to discriminative tasks comes from the lack of inference ability.
To fill this gap, prior work introduces an additional encoder to the GAN structure~\cite{bigan,ali}.
Donahue and Simonyan~\cite{bigbigan} and Pidhorskyi \textit{et al.}~\cite{alae} extend this idea to the state-of-the-art BigGAN~\cite{biggan} and StyleGAN~\cite{stylegan} models respectively.
In this paper, we also study the representation learning using GANs, with following \textbf{improvements} compared to existing methods.
First, we propose to treat the well-trained StyleGAN generator as \textit{a learned loss function}.
Second, instead of mapping the images to the initial GAN latent space, like most algorithms~\cite{bigan,ali,bigbigan,alae} have done, we design a novel encoder to produce \textit{hierarchical} features that well align with the layer-wise representation learned by StyleGAN.
Third, besides the image classification task that is mainly targeted at by prior work~\cite{bigan,ali,bigbigan,alae}, we validate the \textit{transferability} of our proposed GH-Feat on a range of generative and discriminative tasks, demonstrating its generalization ability.

\section{Methodology}\label{sec:method}
We design a novel encoder to extract hierarchical visual features from the input images.
This encoder is trained in an unsupervised learning manner from the image reconstruction loss based on a prepared StyleGAN generator.
Sec.~\ref{subsec:layerwise-representation} describes how we abstract the multi-level representation from StyleGAN.
Sec.~\ref{subsec:hierarchical-encoder} presents the structure of the novel hierarchical encoder.
Sec.~\ref{subsec:stylegan-as-loss} introduces the idea of using pre-trained StyleGAN generator as a learned loss function for representation learning.

\definecolor{myblue}{rgb}{0.6, 0.77, 1.0}
\begin{figure*}[t]
  \centering
  \includegraphics[width=1.0\linewidth]{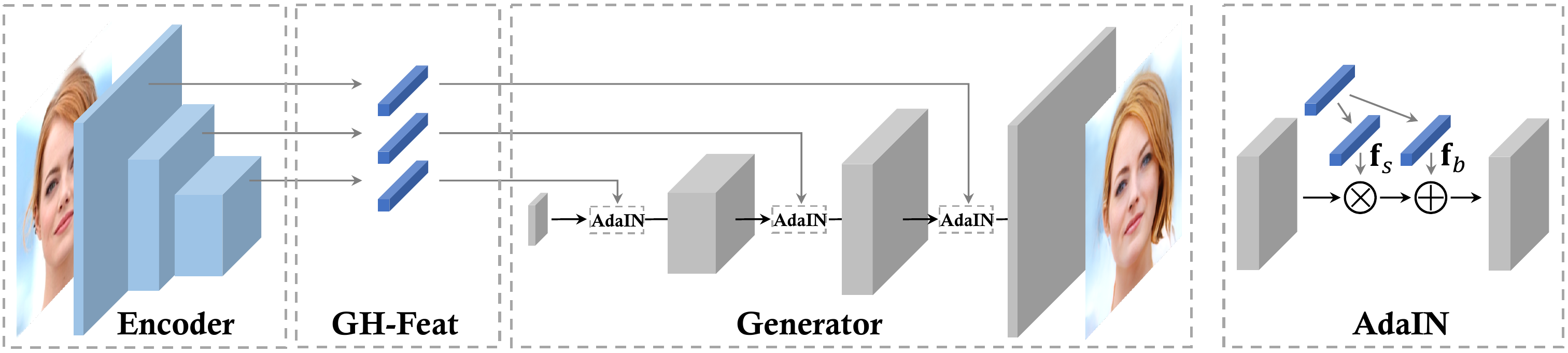}
  \caption{
    Framework of the proposed encoder, which is able to extract \textit{Generative Hierarchical Features (GH-Feat)} from input image.
    This feature hierarchy highly aligns with the layer-wise representation (\textit{i.e.}, style codes of per-layer AdaIN) learned by the StyleGAN generator.
    Parameters in \textbf{\textcolor{myblue}{blue}} blocks are trainable.
  }
  \label{fig:framework}
  \vspace{-10pt}
\end{figure*}

\subsection{Layer-wise Representation from StyleGAN}\label{subsec:layerwise-representation}
\vspace{-1pt}
The generator $G(\cdot)$ of GANs typically takes a latent code $\z\in\Z$ as the input and is trained to synthesize a photo-realistic image $\x=G(\z)$.
The recent state-of-the-art StyleGAN~\cite{stylegan} proposes to first map $\z$ to a disentangled space $\W$ with $\w=f(\z)$.
Here, $f(\cdot)$ denotes the mapping implemented by multi-layer perceptron (MLP).
The $\w$ code is then projected to layer-wise style codes $\{\y^{(\ell)}\}_{\ell=1}^L \triangleq \{(\y_s^{(\ell)}, \y_b^{(\ell)})\}_{\ell=1}^L$ with affine transformations, where $L$ is the number of convolutional layers.
$\y_s^{(\ell)}$ and $\y_b^{(\ell)}$ correspond to the scale and weight parameters in Adaptive Instance Normalization (AdaIN)~\cite{adain}.
These style codes are used to modulate the output feature maps of each convolutional layer with
\begin{align}
  \mathtt{AdaIN}(\x_i^{(\ell)}, \y^{(\ell)}) = \y_{s,i}^{(\ell)}\ \frac{\x_i^{(\ell)} - \mu(\x_i^{(\ell)})}{\sigma(\x_i^{(\ell)})} + \y_{b,i}^{(\ell)}, \label{eq:adain}
\end{align}
where $\x_i^{(\ell)}$ indicates the $i$-th channel of the output feature map from the $\ell$-th layer.
$\mu(\cdot)$ and $\sigma(\cdot)$ denote the mean and variance respectively.

Here, we treat the layer-wise style codes, $\{\y^{(\ell)}\}_{\ell=1}^L$, as the generative visual features that we would like to extract from the input image.
There are two major advantages.
First, the synthesized image can be completely determined by these style codes without any other variations, making them suitable to express the information contained in the input data from the generative perspective.
Second, these style codes are organized as a hierarchy where codes at different layers correspond to semantics at different levels~\cite{stylegan,higan}.
To the best of our knowledge, this is the first work that adopts the style codes for the per-layer AdaIN module as the learned representations of StyleGAN.

\subsection{Hierarchical Encoder}\label{subsec:hierarchical-encoder}
Based on the layer-wise representation described in Sec.~\ref{subsec:layerwise-representation}, we propose a novel encoder $E(\cdot)$ with a hierarchical structure to extract multi-level visual features from a given image.
As shown in Fig.~\ref{fig:framework}, the encoder is designed to best align with the StyleGAN generator.
In particular, the Generative Hierarchical Features (GH-Feat) produced by the encoder, $\{\f^{(\ell)}\}_{\ell=1}^L \triangleq \{(\f_s^{(\ell)}, \f_b^{(\ell)})\}_{\ell=1}^L$, are fed into the per-layer AdaIN module of the generator by replacing the style code $\y^{(L-\ell+1)}$ in Eq.~\eqref{eq:adain}.

We adopt ResNet~\cite{resnet} architecture as the encoder backbone and add an extra residual block to get an additional feature map with lower resolution.%
\footnote{In fact, there are totally six stages in our encoder, where the first one is a convolutional layer (followed by a pooling layer) and each of the others consists of several residual blocks.}
Besides, we introduce a feature pyramid network~\cite{fpn} to learn the features from multiple levels.
The output feature maps from the last three stages, $\{R_4, R_5, R_6\}$, are used to produce GH-Feat.
Taking a 14-layer StyleGAN generator as an instance, $R_4$ aligns with layer 9-14, $R_5$ with 5-8, while $R_6$ with 1-4.
Here, to bridge the feature map with each style code, we first downsample it to $4\times4$ resolution and then map it to a vector of the target dimension using a fully-connect (FC) layer.
In addition, we introduce a lightweight Spatial Alignment Module (SAM)~\cite{tpn, parnet} into the encoder structure to better capture the spatial information from the input image.
SAM works in a simple yet efficient way:
\begin{align}
  R_i &= W_i\mathtt{down}(R_i) + W_6R_6 \quad i \in \{4,5\}, \nonumber
\end{align}
where $W_4$, $W_5$, and $W_6$ (all are implemented with an $1\times1$ convolutional layer) are used to project the feature maps $R_4$, $R_5$, and $R_6$ to have the same number of feature channels respectively.
$R_4$ and $R_5$ are downsampled to the same resolution of $R_6$ before fusion.
The detailed structure of the encoder can be found in \textbf{Appendix}.

\subsection{StyleGAN Generator as Learned Loss}\label{subsec:stylegan-as-loss}
We consider the pre-trained StyleGAN generator as a leaned loss function.
Specifically, we employ a StyleGAN generator to supervise the encoder training with the objective of image reconstruction.
We also introduce a discriminator to compete with the encoder, following the formulation of GANs~\cite{gan}, to ensure the reconstruction quality.
To summarize, the encoder $E(\cdot)$ and the discriminator $D(\cdot)$ are jointly trained with
\begin{align}
  \begin{split}
    \min_{\Theta_E}\mathcal{L}_E =\ &||\x - G(E(\x))||_2 - \lambda_1 \E_{\x}[D(G(E(\x)))] \\
                                    &+ \lambda_2 ||F(\x) - F(G(E(\x)))||_2, \label{eq:encoder}
  \end{split} \\
  \begin{split}
    \min_{\Theta_D}\mathcal{L}_D =\ &\E_{\x}[D(G(E(\x)))] - \E_{\x}[D(\x)] \\
                                    &+ \lambda_3 \E_{\x}[||\nabla_{\x}D(\x)||_2^2], \label{eq:discriminator}
  \end{split}
\end{align}
where $||\cdot||_2$ denotes the $\ell_2$ norm and $\lambda_1, \lambda_2, \lambda_3$ are loss weights to balance different loss terms.
The last term in Eq.~\eqref{eq:encoder} represents the perceptual loss~\cite{johnson2016perceptual} and $F(\cdot)$ denotes the $\mathtt{conv4\_3}$ output from a pre-trained VGG~\cite{vgg} model.

\section{Experiments}\label{sec:experiments}
We evaluate Generative Hierarchical Features (GH-Feat) on a wide range of downstream applications.
Sec.~\ref{exp:setting} introduces the experimental settings, such as implementation details, datasets, and tasks.
Sec.~\ref{exp:ablation} conducts ablation study on the proposed hierarchical encoder.
Sec.~\ref{exp:generative} and Sec.~\ref{exp:discriminative} evaluate the applicability of GH-Feat on generative and discriminative tasks respectively.

\subsection{Experimental Settings}\label{exp:setting}

\noindent\textbf{Implementation Details.}
The loss weights are set as $\lambda_1=0.1$, $\lambda_2=5e^{-5}$, and $\lambda_3=5$.
We use Adam~\cite{adam} optimizer, with $\beta_1=0$ and $\beta_2=0.99$, to train both the encoder and the discriminator.
The learning rate is initially set as $1e^{-4}$ and exponentially decayed with the factor of 0.8.

\vspace{2pt}
\noindent\textbf{Datasets and Models.}
We conduct experiments on four StyleGAN~\cite{stylegan} models, pre-trained on MNIST~\cite{mnist}, FF-HQ~\cite{stylegan}, LSUN bedrooms~\cite{lsun}, and ImageNet~\cite{imagenet} respectively.
The MNIST model is with $32\times32$ resolution and the remaining models are with $256\times256$ resolution.

\vspace{2pt}
\noindent\textbf{Tasks and Metrics.}
Unlike existing adversarial feature learning methods~\cite{ali,bigan,alae,bigbigan} that are mainly evaluated on the high-level image classification task, we benchmark GH-Feat on a range of both generative and discriminative tasks from multiple levels, including
\textit{\textbf{(1)} Image editing.}
It focuses on manipulating the image content or style, \textit{e.g.}, style mixing, global editing, and local editing.
\textit{\textbf{(2)} Image harmonization.}
This task harmonizes a discontinuous image to produce a realistic output.
\textit{\textbf{(3)} MNIST digit recognition.}
It is a long-standing image classification task.
We report the Top-1 accuracy on the test set following~\cite{mnist}.
\textit{\textbf{(4)} Face verification.}
It aims at distinguishing whether the given pair of faces come from the same identity.
We validates on the LFW dataset~\cite{lfw} following the standard protocol~\cite{lfw}.
\textit{\textbf{(5)} ImageNet classification.}
This is a large-scale image classification dataset~\cite{imagenet}, consisting of over $1M$ training samples across 1,000 classes and $50K$ validation samples.
We use Top-1 accuracy as the evaluation metric following existing work~\cite{bigan,bigbigan}.
\textit{\textbf{(6)} Pose estimation.}
This task targets at estimating the yaw pose of the input face.
$70K$ real faces on FF-HQ~\cite{stylegan} are split to $60K$ training samples and $10K$ test samples.
The $\ell_1$ regression error is used as the evaluation metric.
\textit{\textbf{(7)} Landmark detection.}
This task learns a set of semantic points with visual meaning.
We use FF-HQ~\cite{stylegan} dataset and follow the standard MSE metric~\cite{tcdcn,jakab2018unsupervised,xu2020unsupervised} to report performances in inter-ocular distance (IOD).
\textit{\textbf{(8)} Layout prediction.}
We extract the corner points of the layout line and convert the task to a landmark regression task.
The annotations of the collected $90K$ bedroom images ($70K$ for training and $20K$ for validation) are obtained with~\cite{layoutlearinng}.
Following~\cite{zou2018layoutnet}, we report the corner distance as the metric.
\textit{\textbf{(9)} Face luminance regression.}
It focuses on regressing the luminance of face images.
We use it as a low-level task on the FF-HQ~\cite{stylegan} dataset.

\subsection{Ablation Study}\label{exp:ablation}

We make ablation studies on the training of encoder from two perspectives.
(1) We choose the layer-wise style codes $\y$ over the $\w$ codes as the representation from StyleGAN.
(2) We introduce Spatial Alignment Module (SAM) into the encoder to better handle the spatial information.

\setlength{\tabcolsep}{12pt}
\begin{table}[t]
  \caption{
    Quantitative results on ablation study.
  }
  \label{tab:ablation}
  \vspace{2pt}
  \centering\small
  \begin{tabular}{cccccc}
    \toprule
    Space &       SAM &    \MSE & \SSIM &  \FID \\ \midrule
    $\W$  & \ding{51} &  0.0601 & 0.540 & 22.24 \\
    $\Y$  &           &  0.0502 & 0.550 & 19.06 \\
    $\Y$  & \ding{51} &  0.0464 & 0.558 & 18.48 \\ \bottomrule
  \end{tabular}
  \vspace{-5pt}
\end{table}

\setlength{\tabcolsep}{2.5pt}
\begin{table}[t]
  \caption{
    Quantitative comparison with ALAE~\cite{alae} on reconstructing images from FF-HQ faces~\cite{stylegan} and LSUN bedrooms~\cite{lsun}.
  }
  \label{tab:inversion}
  \vspace{2pt}
  \centering\small
  \begin{tabular}{lcccccc}
    \toprule
            & \multicolumn{3}{c}{Face} & \multicolumn{3}{c}{Bedroom} \\ \cmidrule[0.5pt](lr){2-4} \cmidrule[0.5pt](lr){5-7}
    Method           &  \MSE & \SSIM &  \FID &  \MSE & \SSIM &  \FID \\ \midrule
    ALAE~\cite{alae} & 0.182 & 0.398 & 24.86 & 0.275 & 0.315 & 21.01 \\
    GH-Feat (Ours)   & 0.046 & 0.558 & 18.42 & 0.068 & 0.507 & 16.01 \\ \bottomrule
  \end{tabular}
  \vspace{-5pt}
\end{table}

\begin{figure}[!ht]
  \centering
  \includegraphics[width=1.0\linewidth]{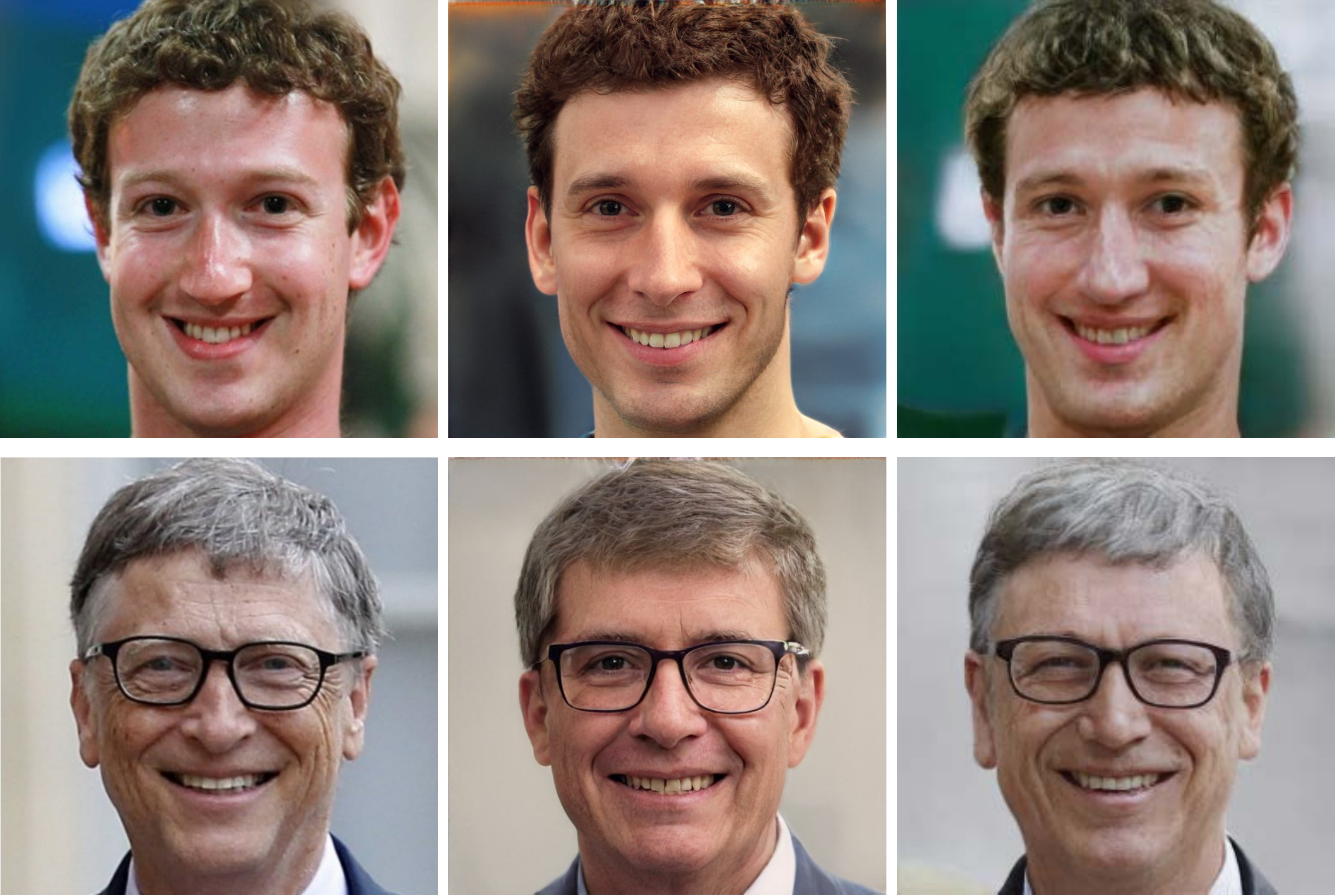}
  \caption{
    Qualitative comparison on reconstructing real images.
    From left to right: Inputs, ALAE~\cite{alae}, and our GH-Feat.
  }
  \label{fig:inversion}
  \vspace{-10pt}
\end{figure}

\begin{figure*}[t]
  \centering
  \includegraphics[width=1.0\linewidth]{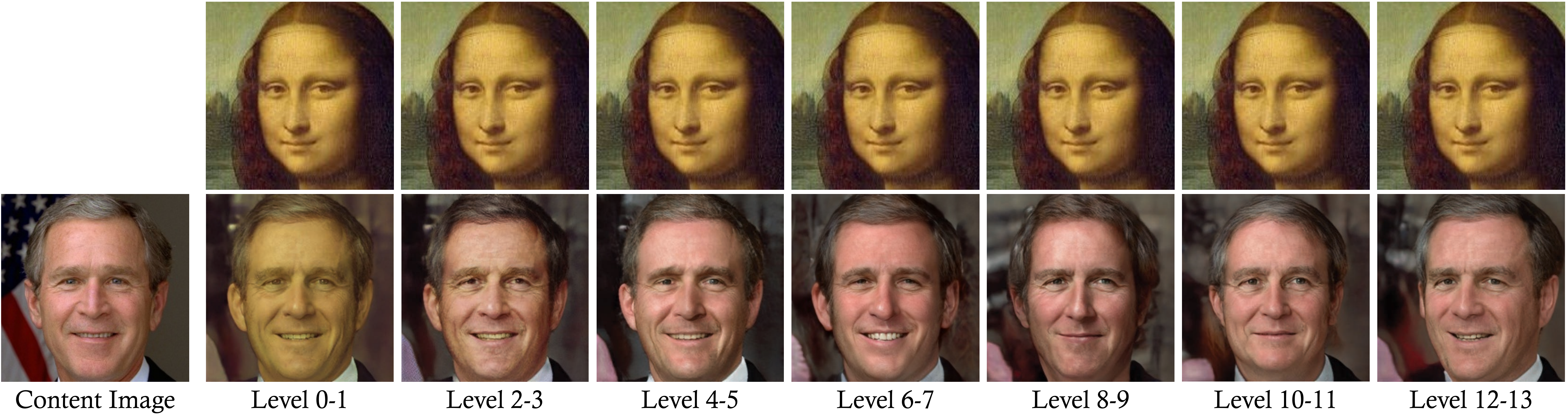}
  \caption{
    \textbf{Style mixing} results by exchanging the GH-Feat extracted from the content image and the style image (first row) at different levels.
    Higher level corresponds to more abstract feature.
  }
  \label{fig:style-mixing}
  \vspace{-10pt}
\end{figure*}

\begin{figure*}[t]
  \centering
  \includegraphics[width=1.0\linewidth]{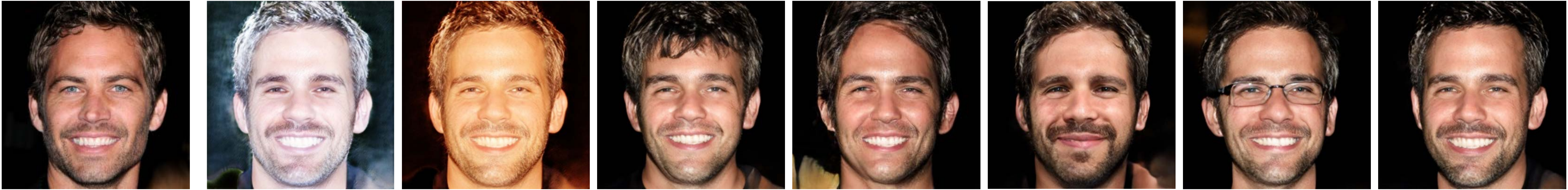}
  \caption{
    \textbf{Global image editing} achieved by GH-Feat.
    On the left is the input image, while the others are generated by randomly sampling the visual feature at some particular level.
  }
  \label{fig:global}
  \vspace{-10pt}
\end{figure*}

Since the encoder is trained with the objective of image reconstruction, we use mean square error (MSE), SSIM~\cite{ssim}, and FID~\cite{fid} to evaluate the encoder performance.
Tab.~\ref{tab:ablation} shows the results where we can tell that our encoder benefits from the effective SAM module and that choosing an adequate representation space (\textit{i.e.}, the comparison between the first row and the last row) results in a better reconstruction.
More discussion on the differences between $\W$ space and $\Y$ space can be found in Sec.~\ref{exp:hierarchical}.

\begin{figure*}[t]
  \centering
  \includegraphics[width=1.0\linewidth]{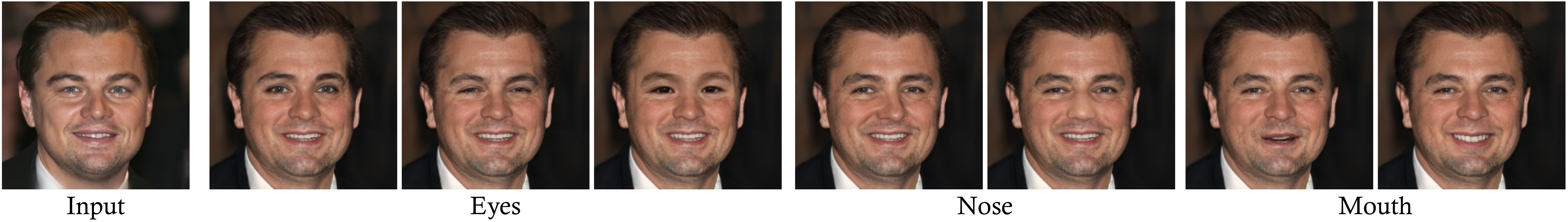}
  \caption{
    \textbf{Local image editing} achieved by GH-Feat.
    On the left is the input image, while the others are generated by randomly sampling the visual feature and replacing the spatial feature map (for different regions) at some particular level.
    Zoom in for details.
  }
  \label{fig:local}
  \vspace{-5pt}
\end{figure*}

\begin{figure*}[t]
  \centering
  \includegraphics[width=1.0\linewidth]{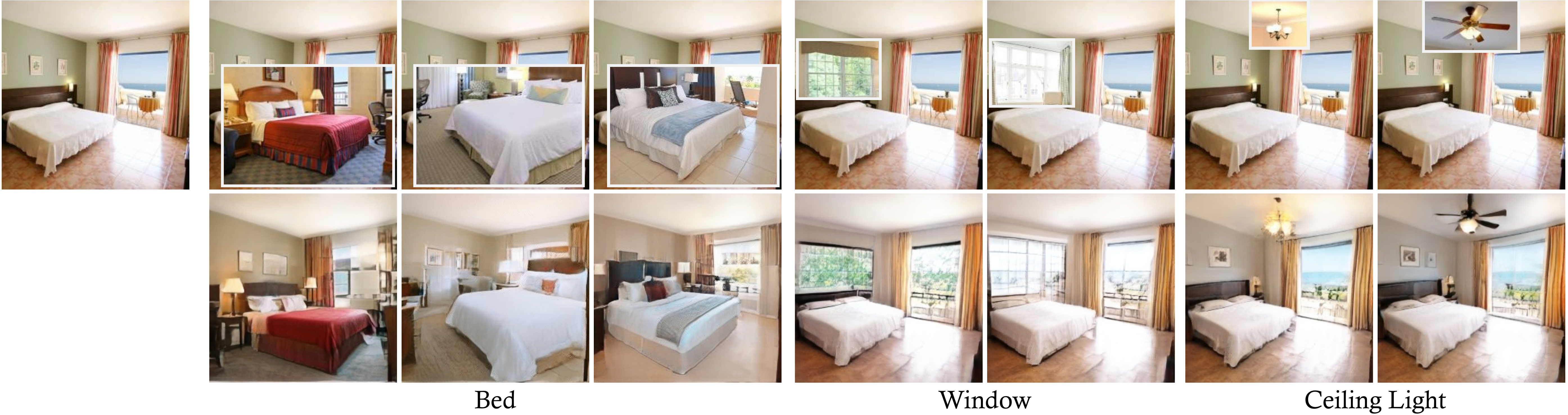}
  \caption{
    \textbf{Image harmonization} with GH-Feat.
    On the top left corner is the original image.
    Pasting a target image patch onto the original image then feeding it as the input (top row), our hierarchical encoder is able to smooth the image content and produce a photo-realistic image (bottom row).
  }
  \label{fig:harmonization}
  \vspace{-5pt}
\end{figure*}

\begin{figure*}[!ht]
  \centering
  \includegraphics[width=1\linewidth]{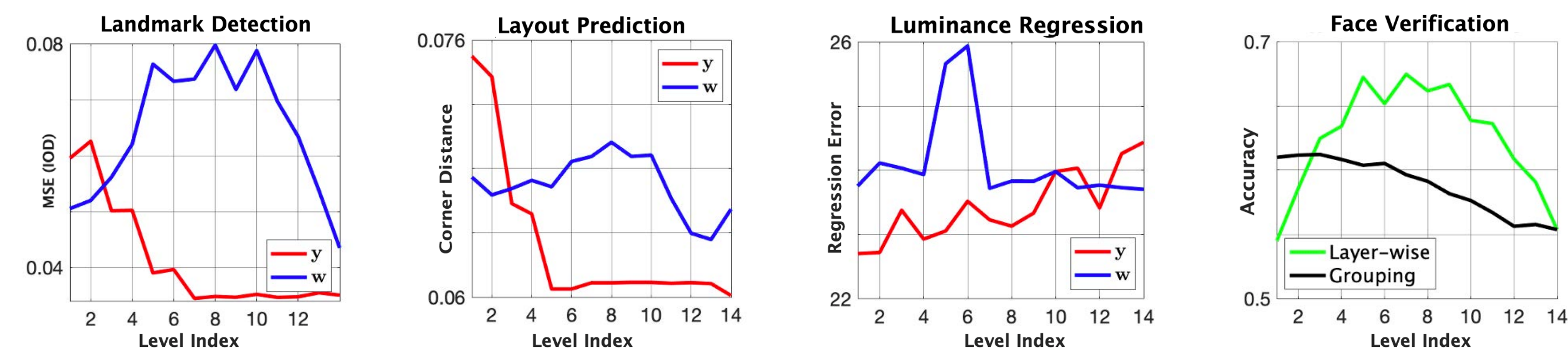}
  \caption{
    Performance on different discriminative tasks using GH-Feat.
    Left three columns enclose the comparisons between using different spaces of StyleGAN as the representation space, where $\Y$ space (in \textbf{\textcolor{red}{red}} color) shows stronger discriminative and hierarchical property than $\W$ space (in \textbf{\textcolor{blue}{blue}} color).
    This is discussed in Sec.~\ref{exp:hierarchical}.
    The last column compares the two different strategies used in the face verification task, which is explained in Sec.~\ref{exp:classification}.
    Higher level corresponds to more abstract feature.
  }
  \label{fig:layerwise}
  \vspace{-5pt}
\end{figure*}

\subsection{Evaluation on Generative Tasks}\label{exp:generative}

Thanks to using the StyleGAN as a learned loss function, a huge advantage of GH-Feat over existing unsupervised feature learning approaches~\cite{hjelm2019learning,zhuang2019local,cpc,cmc,moco}, which mainly focus on the image classification task, is its generative capability.
In this section, we conduct a number of generative experiments to verify this point.

\vspace{-10pt}
\subsubsection{Image Reconstruction}\label{exp:reconstruction}
\vspace{-5pt}
Image reconstruction is an important evaluation on whether the learned features can best represent the input image.
The very recent work ALAE~\cite{alae} also employs StyleGAN for representation learning.
We have following differences from ALAE:
(1) We use the $\Y$ space instead of the $\W$ space of StyleGAN as the representation space.
(2) We learn \textit{hierarchical} features that highly align with the per-layer style codes in StyleGAN.
(3) Our encoder can be \textit{efficiently} trained with a well-learned generator by treating StyleGAN as a loss function.
Tab.~\ref{tab:inversion} and Fig.~\ref{fig:inversion} show the quantitative and qualitative comparison between GH-Feat and ALAE~\cite{alae} on FF-HQ faces~\cite{stylegan} and LSUN bedrooms~\cite{lsun}.
We can tell that GH-Feat better reconstructs the input by preserving more information, resulting a more expressiveness representation.

\vspace{-10pt}
\subsubsection{Image Editing}\label{exp:editing}
\vspace{-5pt}
In this part, we evaluate GH-Feat on a number of image editing tasks.
Different from the features learned from discriminative tasks~\cite{resnet,moco}, our GH-Feat naturally supports sampling and enables creating new data.

\vspace{2pt}
\noindent\textbf{Style Mixing.}
To achieve style mixing, we use the encoder to extract visual features from both the content image and the style image and swap these two features at some particular level.
The swapped features are then visualized by the generator, as shown in Fig.~\ref{fig:style-mixing}.
We can observe the compelling hierarchical property of the learned GH-Feat.
For example, by exchanging low-level features, only the image color tone and the skin color are changed.
Meanwhile, mid-level features controls the expression, age, or even hair styles.
Finally, high-level features correspond to the face shape and pose information (last two columns).

\vspace{2pt}
\noindent\textbf{Global Editing.}
The style mixing results have suggested the potential of GH-Feat in multi-level image stylization.
Sometime, however, we may not have a target style image to use as the reference.
Thanks to the design of the latent space in GANs~\cite{gan}, the generative representation naturally supports sampling, resulting in a strong creativity.
In other words, based on GH-Feat, we can arbitrarily sample meaningful visual features and use them for image editing.
Fig.~\ref{fig:global} presents some high-fidelity editing results at multiple levels.
This benefits from the matching between the learned GH-Feat and the internal representation of StyleGAN.

\vspace{2pt}
\noindent\textbf{Local Editing.}
Besides global editing, our GH-Feat also facilitates editing the target image locally by deeply cooperating with the generator.
In particular, instead of directly swapping features, we can exchange a certain region of the spatial feature map at some certain level.
In this way, only a local patch in the output image will be modified while other parts remain untouched.
As shown in Fig.~\ref{fig:local}, we can successfully manipulate the input face with different eyes, noses, and mouths.

\vspace{-10pt}
\subsubsection{Image Harmonization}\label{exp:harmonize}
\vspace{-5pt}
Our hierarchical encoder is robust such that it can extract reasonable visual features even from discontinuous image content.
We copy some patches (\textit{e.g.}, bed and window) onto a bedroom image and feed the stitched image into our proposed encoder for feature extraction.
The extracted features are then visualized via the generator, as in Fig.~\ref{fig:harmonization}.
We can see that the copied patches well blend into the ``background''.
We also surprisingly find that when copying a window into the source image, the view from the original window and that from the new window highly align with each other (\textit{e.g.}, vegetation or ocean), benefiting from the robust generative visual features.

\subsection{Evaluation on Discriminative Tasks}\label{exp:discriminative}
In this part, we verify that even the proposed GH-Feat is learned from generative models, it can be applicable to a wide range of discriminative tasks with competitive performances.
Here, we do not fine-tune the encoder for any certain task.
In particular, we choose multi-level downstream applications, including image classification, face verification, pose estimation, layout prediction, landmark detection, and luminance regression.
For each task, we use our encoder to extract visual features from both the training and the test set.
A linear regression model (\textit{i.e.}, a fully-connected layer) is learned on the training set with ground-truth and then evaluated on the test set.

\begin{figure*}[t]
  \centering
  \includegraphics[width=1.0\linewidth]{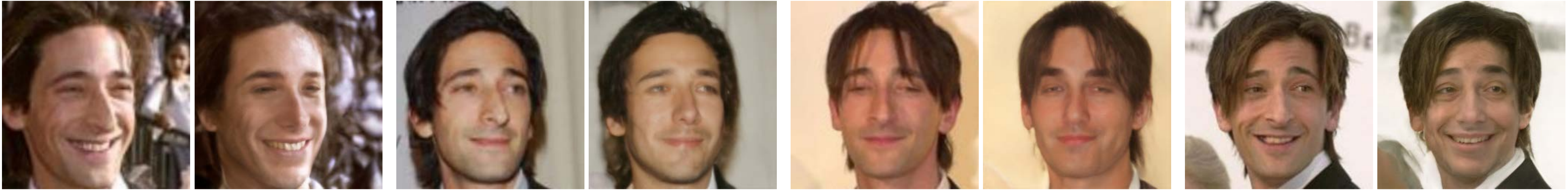}
  \caption{
    \textbf{Image reconstruction} results on LFW~\cite{lfw}.
    For each pair of images, left is the low-resolution input while right is reconstructed by GH-Feat.
    All samples are with the same identity.
  }
  \label{fig:lfw}
  \vspace{-10pt}
\end{figure*}

\vspace{-10pt}
\subsubsection{Discriminative and Hierarchical Property}\label{exp:hierarchical}
\vspace{-5pt}
Recall that GH-Feat is a multi-scale representation learned by using StyleGAN as a loss function.
As a results, it consists of features from multiple levels, each of which correspond to a certain layer in the StyleGAN generator.
Here, we would to explore how this feature hierarchy is organized as well as how they can facilitate multi-level discriminative tasks, including face pose estimation, indoor scene layout prediction, and luminance\footnote{We convert images from RGB space to YUV space and use the mean value from Y space as the luminance.} regression from face images.
In particular, we evaluate GH-Feat on each task level by level.
As a comparison, we also train encoders by treating the $\w$ code, instead of the style code $\y$, as the representation.
From Fig.~\ref{fig:layerwise}, we have three observations:
(1) GH-Feat is discriminative.
(2) Features at lower level are more suitable for low-level tasks (\textit{e.g.}, luminance regression) and those at higher level better aid high-level tasks (\textit{e.g.}, pose estimation).
(3) $\Y$ space demonstrates a more obvious hierarchical property than $\W$ space.

\vspace{-10pt}
\subsubsection{Digit Recognition \& Face Verification}\label{exp:classification}
\vspace{-5pt}
Image classification is widely used to evaluate the performance of learned representations~\cite{hjelm2019learning,zhuang2019local,moco,cpc,bigbigan}.
In this section, we first compare our proposed GH-Feat with other alternatives on a toy dataset, \textit{i.e.}, MNIST \cite{mnist}.
Then, we use a more challenging task, \textit{i.e.}, face verification, to evaluate the discriminative property of GH-Feat.

\vspace{2pt}
\noindent\textbf{MNIST Digit Recognition.}
We first show a toy example on MNIST following prior work~\cite{bigan,alae}.
We make a little modification to ResNet-18 like~\cite{pytorch-cifar10} which is widely used in literatures to handle samples from MNIST~\cite{mnist} in lower resolution.
The Top-1 accuracy is reported in Tab.~\ref{tab:comparison}~(a).
Our GH-feat outperforms ALAE~\cite{alae} and BiGAN~\cite{bigan} with $1.45\%$ and $1.92\%$, suggesting a stronger discriminative power.
Here, ResNet-18~\cite{resnet} is employed as the backbone structure for both MoCo~\cite{moco} and GH-Feat.

\vspace{2pt}
\noindent\textbf{LFW Face Verification.}
We directly use the proposed encoder, which is trained on FF-HQ~\cite{stylegan}, to extract GH-Feat from face images in LFW~\cite{lfw} and tries three different strategies on exploiting GH-Feat for face verification:
(1) using a single level feature;
(2) grouping multi-level features (starting from the highest level) together;
(3) voting by choosing the largest face similarity across all levels.
Fig.~\ref{fig:layerwise} (last column) shows the results from the first two strategies.
Obviously, GH-Feat from the 5-th to the 9-th levels best preserve the identity information.
Tab.~\ref{tab:comparison}~(b) compares GH-Feat with other unsupervised feature learning methods, including VAE~\cite{vae}, MoCo~\cite{moco}, and ALAE~\cite{alae}.
All these competitors are also trained on FF-HQ dataset~\cite{stylegan} with optimally chosen hyper-parameters.
ResNet-50~\cite{resnet} is employed as the backbone for MoCo and GH-Feat.
Our method with voting strategy achieves 69.7\% accuracy, surpassing other competitors by a large margin.
We also visualize some reconstructed LFW faces in Fig.~\ref{fig:lfw}, where our GH-Feat well handles the domain gap (\textit{e.g.}, image resolution) and preserves the identity information.

\begin{table}[t]
  \caption{
    Quantitative comparison between our proposed GH-Feat and other alternatives on MNIST~\cite{mnist} and LFW~\cite{lfw}.
  }
  \label{tab:comparison}
  \vspace{2pt}
  \centering
  \subfigure[Digit recognition on MNIST.]{
    \centering\small
    \setlength{\tabcolsep}{5.5pt}
    \begin{tabular}{lc}
      \toprule
      Methods                           &  Acc. \\ \midrule
      AE($\ell_{1}$)~\cite{autoencoder} & 97.43 \\
      AE($\ell_{2}$)~\cite{autoencoder} & 97.37 \\
      BiGAN~\cite{bigan}                & 97.14 \\
      ALAE~\cite{alae}                  & 97.61 \\
      MoCo-R18~\cite{moco}              & 95.89      \\ \midrule
      GH-Feat (Ours)                    & 99.06 \\ \bottomrule
    \end{tabular}
  }
  \hspace{2pt}
  \subfigure[Face verification on LFW.]{
    \centering\small
    \setlength{\tabcolsep}{4.5pt}
    \begin{tabular}{lc}
      \toprule
      Methods              & Acc. \\ \midrule
      VAE~\cite{vae}       & 49.3 \\
      MoCo-R50~\cite{moco} & 48.9 \\
      ALAE~\cite{alae}     & 55.7 \\ \midrule
      GH-Feat (Grouping)   & 60.1 \\
      GH-Feat (Layer-wise) & 67.5 \\
      GH-Feat (Voting)     & 69.7 \\ \bottomrule
    \end{tabular}
  }
  \vspace{-10pt}
\end{table}

\setlength{\tabcolsep}{8pt}
\begin{table}[t]
  \caption{
    Quantitative comparison on the ImageNet~\cite{imagenet} classification task.
  }
  \label{tab:imagenet_cls}
  \vspace{2pt}
  \centering\small
  \begin{tabular}{llc}
    \toprule
    Method                                   & Architecture & Top-1 Acc. \\ \midrule
    Motion Seg (MS)~\cite{motionseg,carl}    &   ResNet-101 &       27.6 \\
    Exemplar (Ex)~\cite{exemplar,carl}       &   ResNet-101 &       31.5 \\
    Relative Po (RP)~\cite{relativepos,carl} &   ResNet-101 &       36.2 \\
    Colorization (Col)~\cite{colorful,carl}  &   ResNet-101 &       39.6 \\ \midrule
    \multicolumn{3}{c}{\textit{Contrastive Learning}} \vspace{2pt}       \\
    InstDisc~\cite{instdisc}                 &    ResNet-50 &       42.5 \\
    CPC~\cite{cpc}                           &   ResNet-101 &       48.7 \\
    MoCo~\cite{moco}                         &    ResNet-50 &       60.6 \\ \midrule
    \multicolumn{3}{c}{\textit{Generative Modeling}} \vspace{2pt}        \\
    BiGAN~\cite{bigan}                       &      AlexNet &       31.0 \\
    SS-GAN~\cite{ssgan}                      &    ResNet-19 &       38.3 \\
    BigBiGAN~\cite{bigbigan}                 &    ResNet-50 &       55.4 \\ \midrule
    GH-Feat (Ours)                           &    ResNet-50 &       51.1 \\ \bottomrule
  \end{tabular}
  \vspace{-10pt}
\end{table}

\begin{figure*}[t]
  \centering
  \includegraphics[width=1.0\linewidth]{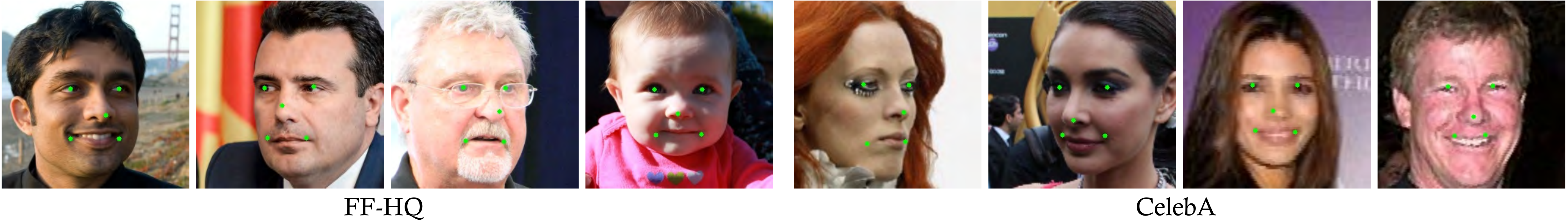}
  \caption{
    \textbf{Landmark detection} results.
    GH-Feat is trained on FF-HQ~\cite{stylegan} dataset but can successfully handle the hard cases (large pose and low image quality) in MAFL dataset~\cite{tcdcn}, a subset of CelebA~\cite{celeba}.
  }
  \label{fig:landmark}
  \vspace{-5pt}
\end{figure*}

\begin{figure*}[t]
  \centering
  \includegraphics[width=1.0\linewidth]{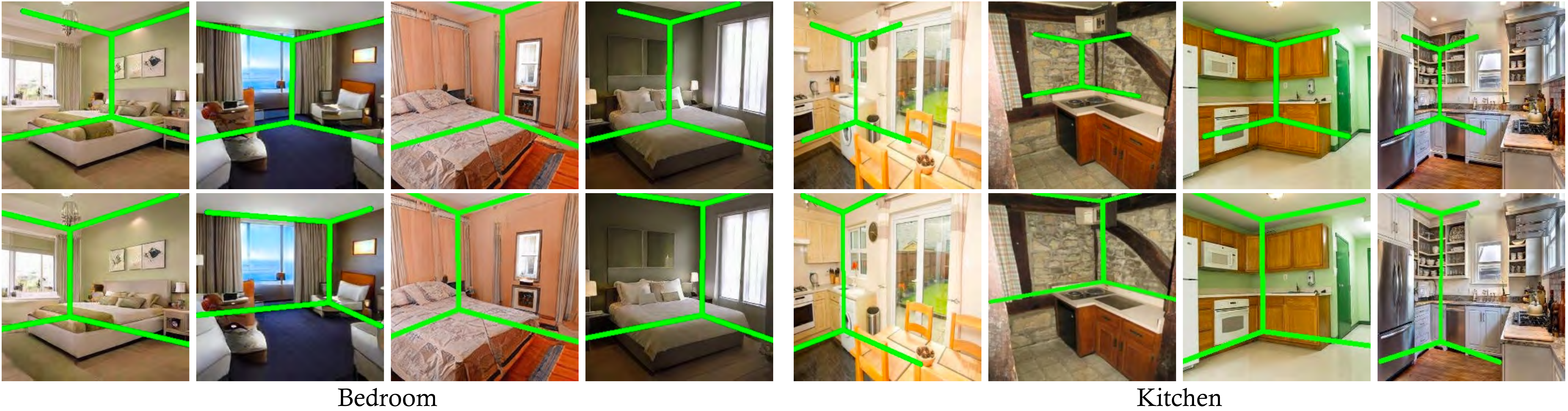}
  \caption{
    \textbf{Layout prediction} results using feature learned by MoCo~\cite{moco} (top row) and our GH-Feat (bottom row).
    Both methods are trained on LSUN bedrooms~\cite{lsun} and then transferred to LSUN kitchens.
  }
  \label{fig:layout}
  \vspace{-5pt}
\end{figure*}

\vspace{-10pt}
\subsubsection{Large-Scale Image Classification}
\vspace{-5pt}
We further evaluate GH-Feat on the high-level image classification task using ImageNet~\cite{imagenet}.
Before the training of encoder, we first train a StyleGAN model, with $256\times256$ resolution, on the ImageNet training collection.
After that, we learn the hierarchical encoder by using the pre-trained generator as the supervision.
No labels are involved in the above training process.%
\footnote{Our encoder can be trained very efficiently, usually $3\times$ faster than the GAN training.}
For the image classification problem, we train a linear model on top of the features extracted from the training set with the softmax loss.
Then, this linear model is evaluated on the validation set.%
\footnote{During testing, we adopt the fully convolutional form as in \cite{googlenet} and average the scores at multiple scales.}
Tab.~\ref{tab:imagenet_cls} shows the comparison between GH-Feat and other unsupervised representation learning approaches~\cite{instdisc,cpc,moco,bigan,ssgan,bigbigan}, where we beat most of the competitors.
The state-of-the-art MoCo~\cite{moco} gives the most compelling performance.
But different from the representations learned with contrastive learning, GH-Feat has huge advantages in generative tasks, as already discussed in Sec.~\ref{exp:generative}.
Among adversarial representation learning approaches, BigBiGAN~\cite{bigbigan} achieves the best performance, benefiting from the incredible large-scale training.
However, GH-Feat presents a stronger generative ability, suggested by the comparison results on image reconstruction shown in Tab.~\ref{tab:image_rec}.
More discussion can be found in \textbf{Appendix}.

\vspace{-10pt}
\subsubsection{Transfer Learning}\label{exp:transfer}
\vspace{-5pt}
In this part, we explore how GH-Feat can be transferred from one dataset to another.

\vspace{2pt}
\noindent\textbf{Landmark Detection.}
We train a linear regression model using GH-Feat on FF-HQ~\cite{stylegan} and test it on MAFL~\cite{tcdcn}, which is a subset of CelebA~\cite{celeba}.
This two datasets have a large domain gap, \textit{e.g.}, faces in MAFL have larger poses yet lower image quality.
As shown in Fig.~\ref{fig:landmark}, GH-Feat shows a strong transferability across these two datasets.
We compare our approach with some supervised and unsupervised alternatives~\cite{tcdcn,mtcnn,jakab2018unsupervised,moco}.
For a fair comparison, we try the multi-scale representations from MoCo~\cite{moco} (\textit{i.e.}, Res2, Res3, Res4, and Res5 feature maps) and report the best results.
Tab.~\ref{tab:landmark} demonstrates the strong generalization ability of GH-Feat.
In particular, it achieves on-par or better performance than the methods that are particular designed for this task~\cite{tcdcn,mtcnn,jakab2018unsupervised}.
Also, it outperforms MoCo~\cite{moco} on this mid-level discriminative task.

\vspace{2pt}
\noindent\textbf{Layout Prediction.}
We train the layout predictor on LSUN~\cite{lsun} bedrooms and test it on kitchens to validate how GH-Feat can be transferred from one scene category to another.
Feature learned by MoCo~\cite{moco} on the bedroom dataset is used for comparison.
We can tell from Fig.~\ref{fig:layout} that GH-Feat shows better predictions than MoCo, especially on the target set (\textit{i.e.}, kitchens), suggesting a stronger transferability.
Like landmark detection, we also conduct experiments with the 4-level representations from MoCo~\cite{moco} and select the best.

\setlength{\tabcolsep}{13pt}
\begin{table}[t]
  \caption{
    Qualitative comparison between BigBiGAN~\cite{bigbigan} and GH-Feat on reconstructing images from ImageNet~\cite{imagenet}.
  }
  \label{tab:image_rec}
  \vspace{2pt}
  \centering\small
  \begin{tabular}{lccc}
    \toprule
                             &  \MSE & \SSIM &  \FID \\ \midrule
    BigBiGAN~\cite{bigbigan} & 0.363 & 0.236 & 33.42 \\
    GH-Feat (Ours)           & 0.078 & 0.431 & 22.70 \\ \bottomrule
  \end{tabular}
  \vspace{-5pt}
\end{table}

\setlength{\tabcolsep}{15pt}
\begin{table}[t]
  \caption{Landmark detection results on MAFL~\cite{tcdcn}.}
  \label{tab:landmark}
  \vspace{2pt}
  \centering\small
  \begin{tabular}{lcc}
    \toprule
    Method                                   & Supervision &  \MSE \\ \midrule
    TCDCN~\cite{tcdcn}                       &   \ding{51} &  7.95 \\
    MTCNN~\cite{mtcnn}                       &   \ding{51} &  5.39 \\
    Cond. ImGen~\cite{jakab2018unsupervised} &             &  4.95 \\
    ALAE~\cite{alae}.                        &             & 10.13 \\
    MoCo-R50~\cite{moco}                     &             &  9.07 \\ \midrule
    GH-Feat (Ours)                           &             &  5.12 \\ \bottomrule
  \end{tabular}
  \vspace{-10pt}
\end{table}

\vspace{-1pt}
\section{Conclusion}\label{sec:conclusion}
\vspace{-2pt}
In this work, we consider the well-trained GAN generator as a learned loss function for learning multi-scale features.
The resulting Generative Hierarchical Features are shown to be generalizable to a wide range of vision tasks.

\vspace{2pt}
\noindent{\textbf{Acknowledgements:}} This work is supported in part by the Early Career Scheme (ECS) through the Research Grants Council (RGC) of Hong Kong under Grant No.24206219, CUHK FoE RSFS Grant, SenseTime Collaborative Grant, and Centre for Perceptual and Interactive Intelligence (CPII) Ltd under the Innovation and Technology Fund.
\setlength{\tabcolsep}{8pt}
\begin{table*}[t]
  \caption{
    \textbf{Encoder Structure}, which is based on ResNet-50~\cite{resnet}.
    Fully-connected (FC) layers are employed to map the feature maps produced by the Spatial Alignment Module (SAM) to our proposed Generative Hierarchical Features (GH-Feat).
    GH-Feat exactly align with the multi-scale style codes used in StyleGAN~\cite{stylegan}.
    The numbers in brackets indicate the dimension of features at each level.
  }
  \label{tab:arch}
  \vspace{2pt}
  \centering\small
  \begin{tabular}{ccccccc}
    \toprule
    Stage & Encoder Pathway &   Output Size & SAM  & FC Dimension & GH-Feat & Style Code in StyleGAN \\
    \midrule
    \multirow{2}{*}{input} & \multirow{2}{*}{$-$} &  \multirow{2}{*}{$3\times 256^2$} & & & & \\
    & & & & & & \\
    \midrule
    \multirow{2}{*}{conv$_1$} & \multicolumn{1}{c}{7$\times$7, {64}} & \multirow{2}{*}{$64\times 128^2$} & & & & \\
    & stride 2, 2 & & & & &\\
    \midrule
    \multirow{2}{*}{pool$_1$} & \multicolumn{1}{c}{3$\times$3, max} &  \multirow{2}{*}{$64\times 64^2$} & & & & \\
    & stride 2, 2 & & & & &\\
    \midrule
    \multirow{3}{*}{res$_2$} & \blocks{{256}}{{64}}{3} & \multirow{3}{*}{$256\times 64^2$} & & & & \\
    & & & & & & \\
    & & & & & & \\
    \midrule
    \multirow{3}{*}{res$_3$} & \blocks{{512}}{{128}}{4}  &  \multirow{3}{*}{$512\times 32^2$} & & & &  \\
    & & & & & & \\
    & & & & & & \\
    \midrule
    \multirow{3}{*}{res$_4$} & \blocks{{1024}}{{256}}{6} &  \multirow{3}{*}{$1024\times 16^2$} & \multirow{3}{*}{$512\times 4^2$} &\multirow{3}{*}{8192$\times$1792} & Level 1-2 & Layer 14-13 ($128d\times2$) \\
    & & & & & Level 3-4  & Layer 12-11 ($256d\times2$) \\
    & & & & & Level 5-6  & Layer 10-9 ($512d\times2$) \\
    \midrule
    \multirow{3}{*}{res$_5$} & \blocks{{2048}}{{512}}{3} & \multirow{3}{*}{$2048\times 8^2$} &  \multirow{3}{*}{$512\times 4^2$} &\multirow{3}{*}{8192$\times$4096} & Level 7-8  & Layer 8-7 ($1024d\times2$) \\
    & & & & & Level 9-10  & Layer 6-5 ($1024d\times2$) \\
    & & & & & & \\
    \midrule
    \multirow{3}{*}{res$_6$} & \blocks{{2048}}{{512}}{1} & \multirow{3}{*}{$2048\times 4^2$} &  \multirow{3}{*}{$512\times 4^2$} &\multirow{3}{*}{8192$\times$4096} & Level 11-12  & Layer 4-3 ($1024d\times2$) \\
    & & & & & Level 13-14  & Layer 2-1 ($1024d\times2$) \\
    & & & & & & \\
    \bottomrule
  \end{tabular}
  \vspace{-10pt}
\end{table*}

\appendix
\section*{Appendix}

\section{Encoder Structure}\label{sec:structure}
Tab.~\ref{tab:arch} provides the detailed architecture of our hierarchical encoder by taking a 14-layer StyleGAN~\cite{stylegan} generator as an instance.
The design of GH-Feat treats the layer-wise style codes used in the StyleGAN model (\textit{i.e.}, the code fed into the AdaIN module~\cite{adain}) as generative features.
Accordingly, GH-Feat consists of 14 levels that exactly align with the multi-scale style codes yet in a reverse order, as shown in the last two columns of Tab.~\ref{tab:arch}.
In particular, these features are projected from the feature maps produced by the last three stages via fully-connected layers.

\begin{figure*}[t]
  \centering
  \includegraphics[width=1\linewidth]{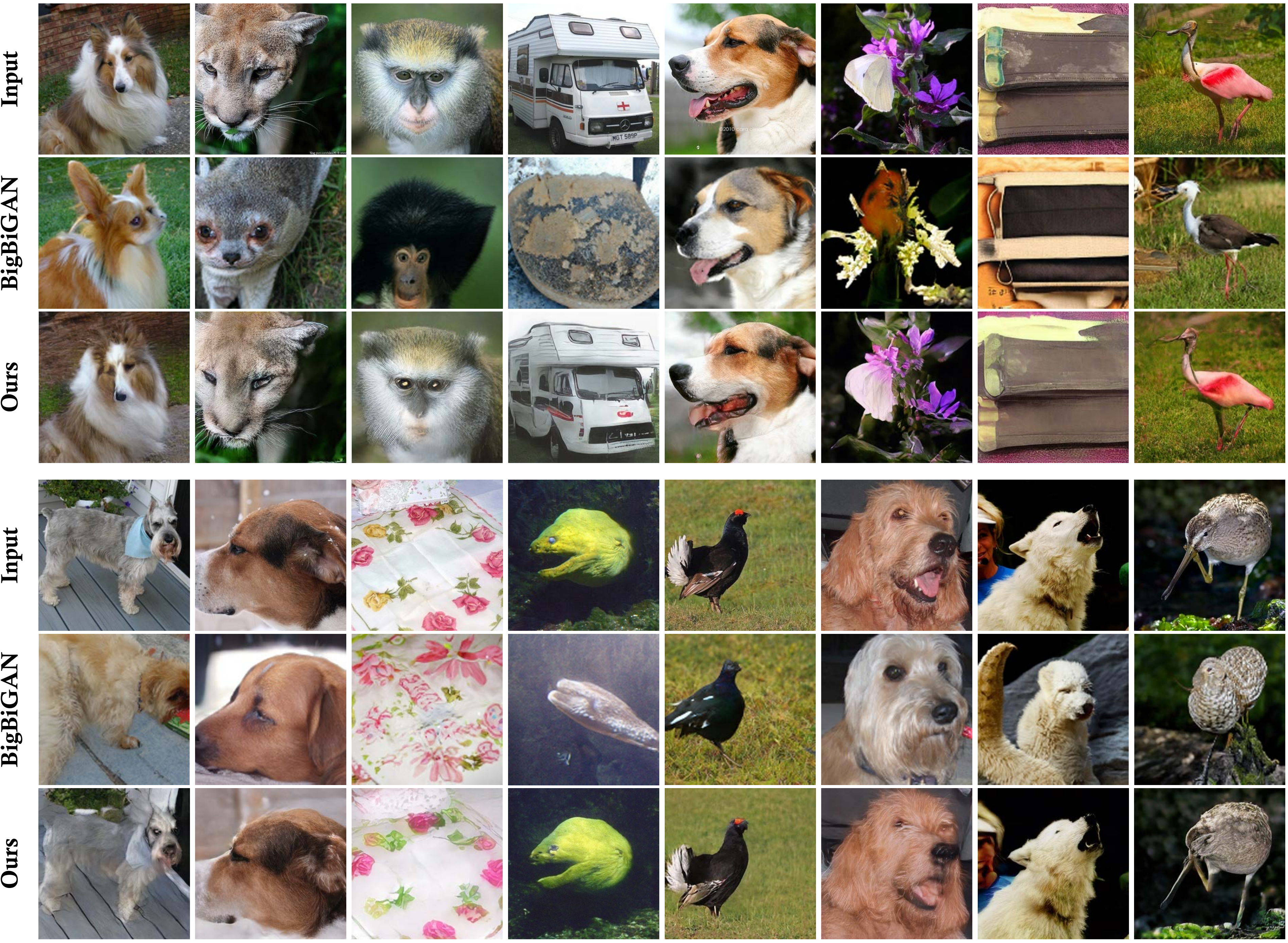}
  \caption{
    Qualitative comparison between BigBiGAN~\cite{bigbigan} and GH-Feat on reconstructing images from ImageNet~\cite{imagenet}.
  }
  \label{fig:imagenet}
  \vspace{-5pt}
\end{figure*}

\section{More Details and Results on ImageNet} \label{sec:imagenet}

\noindent{\textbf{Training Details.}}
During the training of the StyleGAN model on the ImageNet dataset~\cite{imagenet}, we resize all images in the training set such that the short side of each image is 256, and then centrally crop them to $256\times256$ resolution.
All training settings follow the StyleGAN official implementation~\cite{stylegan_github}, including the progressive strategy, optimizer, learning rate, \textit{etc}.
The generator and the discriminator are alternatively optimized until the discriminator have seen $250M$ real images.
After that, the generator is fixed and treated as a well-learned loss function to guide the training of the encoder.
During the training of the hierarchical encoder, images in the training collection are pre-processed in the same way as mentioned above.
After the encoder is ready (usually trained for 25 epochs), we treat it as a feature extractor.
We use the output feature map at the ``res$_5$'' stage (with dimension $2048\times8^2$), apply adaptively average pooling to obtain $2\times2$ spatial feature, and vectorize it.
A linear classifier, \textit{i.e.}, with one fully-connected layer, takes these extracted features as the inputs to learn the image classification task.
SGD optimizer, together with batch size 2048, is used.
The learning rate is initially set as 1 and decayed to 0.1 and 0.01 at the 60-th and the 80-th epoch respectively.
During the training of the final classifier, ResNet-style data augmentation~\cite{resnet} is applied.

\vspace{2pt}
\noindent{\textbf{Discussion.}}
As shown in the main paper, GH-Feat achieves comparable accuracy to existing alternatives.
Especially, among all of methods based on generative modeling, GH-Feat obtains second performance only to BigBiGAN~\cite{bigbigan}, which requires incredible large-scale training.%
\footnote{As reported in \cite{biggan}, the model train on images of $256\times256$ resolution requires 256 TPUs running for 48 hours.}
However, our GH-Feat facilitates a wide rage of tasks besides image classification.
Taking image reconstruction as an example, our approach can well recover the input image, significantly outperforming BigBiGAN~\cite{bigbigan}.
As shown in Fig.~\ref{fig:imagenet}, BigBiGAN can only reconstruct the input image from the category level (\textit{i.e.}, dog or bird).
By contrast, GH-Feat is able to recover more details, like shape and texture.

\setlength{\tabcolsep}{8pt}
\begin{table}[t]
  \caption{
    Quantitative comparison on image reconstruction between training the generator from scratch together with the encoder and our GH-Feat that treats the well-learned StyleGAN generator as a loss function.
  }
  \label{tab:random_g}
  \vspace{2pt}
  \centering\small
  \begin{tabular}{lccc}
    \toprule
                                     &   \MSE & \SSIM &  \FID \\ \midrule
    Training $G(\cdot)$ from Scratch &  0.429 & 0.301 & 46.20 \\
    GH-Feat (Ours)                   & 0.0464 & 0.558 & 18.48 \\ \bottomrule
  \end{tabular}
  \vspace{-10pt}
\end{table}

\section{Ablation Study}\label{sec:ablation}
\begin{figure}[t]
  \centering
  \includegraphics[width=1.0\linewidth]{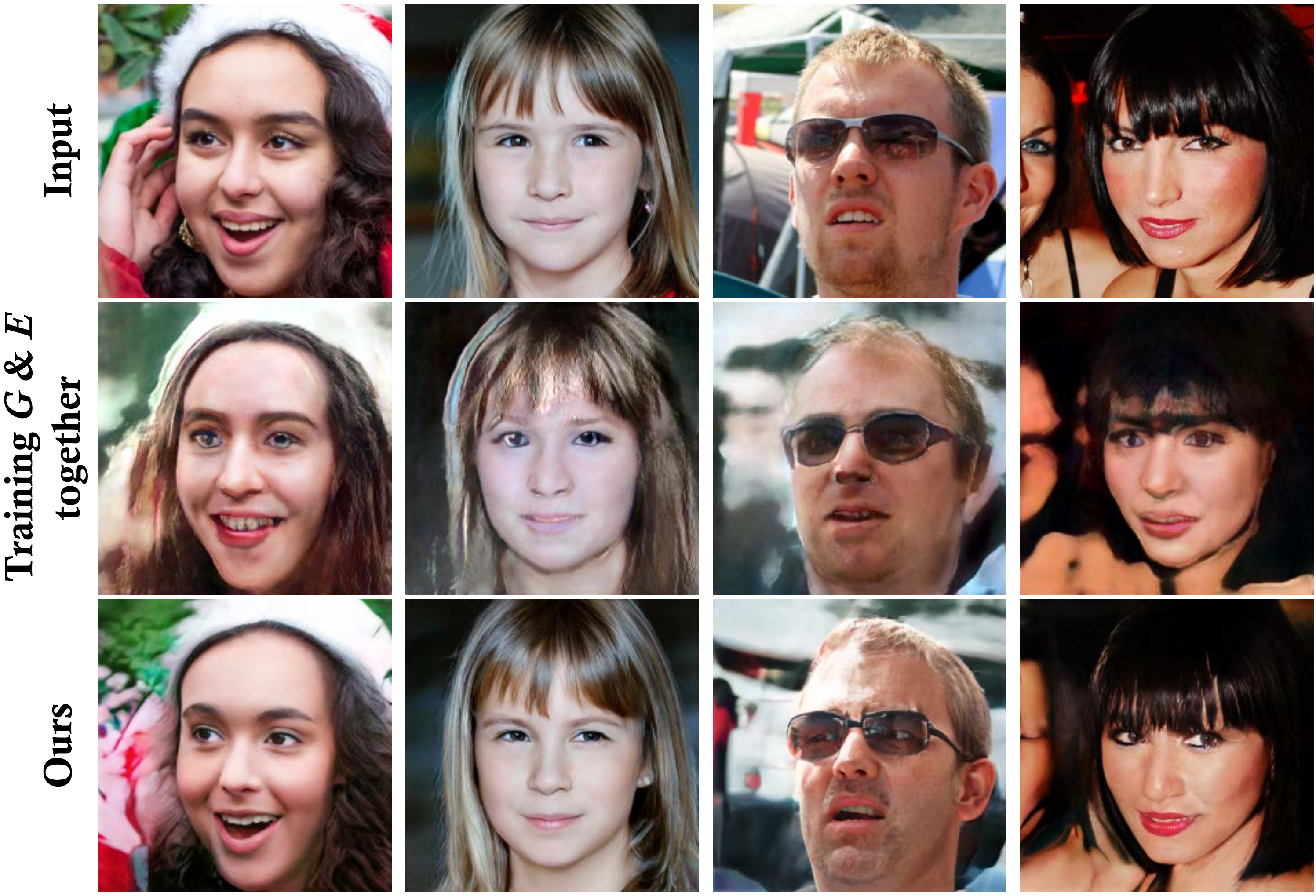}
  \caption{
    Qualitative comparison on image reconstruction between training the generator from scratch together with the encoder, and our GH-Feat that treats the well-learned StyleGAN generator as a loss function.
  }
  \label{fig:random_g}
  \vspace{-10pt}
\end{figure}

Recall that, during the training of the encoder, we propose to treat the well-trained StyleGAN generator as a learned loss function.
In this part, we explore what will happen if we train the generator from scratch together with the encoder.
Tab.~\ref{tab:random_g} and Fig.~\ref{fig:random_g} show the quantitative and qualitative results respectively, which demonstrate the strong performance of GH-Feat.
It suggests that besides higher efficiency, reusing the knowledge from a well-trained generator can also bring better performance.

\begin{figure*}[t]
  \centering
  \includegraphics[width=1.0\linewidth]{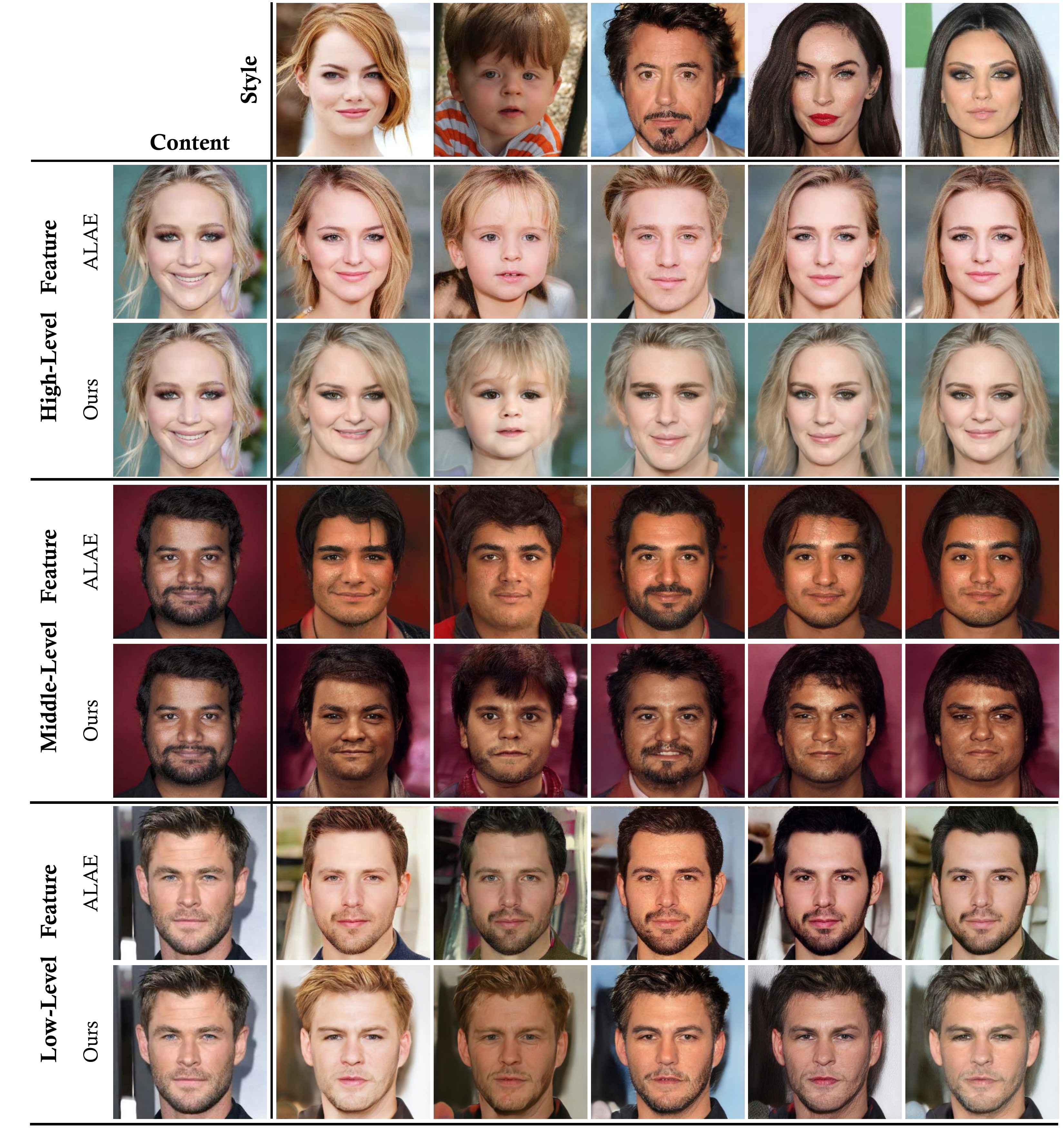}
  \caption{
    Qualitative comparison between our proposed GH-Feat and ALAE~\cite{alae} on the style mixing task.
    After extracting features from both content images and style images, we replace different levels of features from content images with those from style images.
  }
  \label{fig:stylemix}
  \vspace{0pt}
\end{figure*}

\section{Style Mixing}\label{sec:mixing}
In this part, we verify the hierarchical property of GH-Feat on the task of style mixing and further make comparison with ALAE~\cite{alae}.
In particular, we use ALAE and our approach to extract features from same images (including both style images and content images) and then use the extracted features for style mixing at different levels.

Fig.~\ref{fig:stylemix} shows the comparison results.
Note that all test images are selected following the original paper of ALAE~\cite{alae}.
We can see that when mixing high-level styles, the pose, age, and gender of mixed results are close to those of style images.
By comparing with ALAE, results using GH-Feat better preserve the identity information (high-level feature) from style images as well as the color information (low-level feature) from content images.
In addition, when mixing low-level styles (bottom two rows), both ALAE and GH-Feat can successfully transfer the color style from style images to content images, but GH-Feat shows much stronger identity preservation.

{\small
\bibliographystyle{ieee_fullname}
\bibliography{ref}

\begin{thebibliography}{10}\itemsep=-1pt

\bibitem{bau2017network}
David Bau, Bolei Zhou, Aditya Khosla, Aude Oliva, and Antonio Torralba.
\newblock Network dissection: Quantifying interpretability of deep visual
  representations.
\newblock In {\em IEEE Conf. Comput. Vis. Pattern Recog.}, 2017.

\bibitem{gandissect}
David Bau, Jun-Yan Zhu, Hendrik Strobelt, Bolei Zhou, Joshua~B. Tenenbaum,
  William~T. Freeman, and Antonio Torralba.
\newblock Gan dissection: Visualizing and understanding generative adversarial
  networks.
\newblock In {\em Int. Conf. Learn. Represent.}, 2019.

\bibitem{surf}
Herbert Bay, Tinne Tuytelaars, and Luc Van~Gool.
\newblock Surf: Speeded up robust features.
\newblock In {\em Eur. Conf. Comput. Vis.}, 2006.

\bibitem{bengio2013representation}
Yoshua Bengio, Aaron Courville, and Pascal Vincent.
\newblock Representation learning: A review and new perspectives.
\newblock {\em IEEE Trans. Pattern Anal. Mach. Intell.}, 2013.

\bibitem{biggan}
Andrew Brock, Jeff Donahue, and Karen Simonyan.
\newblock Large scale {GAN} training for high fidelity natural image synthesis.
\newblock In {\em Int. Conf. Learn. Represent.}, 2019.

\bibitem{ssgan}
Ting Chen, Xiaohua Zhai, Marvin Ritter, Mario Lucic, and Neil Houlsby.
\newblock Self-supervised gans via auxiliary rotation loss.
\newblock In {\em IEEE Conf. Comput. Vis. Pattern Recog.}, 2019.

\bibitem{hog}
Navneet Dalal and Bill Triggs.
\newblock Histograms of oriented gradients for human detection.
\newblock In {\em IEEE Conf. Comput. Vis. Pattern Recog.}, 2005.

\bibitem{imagenet}
Jia Deng, Wei Dong, Richard Socher, Li-Jia Li, Kai Li, and Li Fei-Fei.
\newblock Imagenet: A large-scale hierarchical image database.
\newblock In {\em IEEE Conf. Comput. Vis. Pattern Recog.}, 2009.

\bibitem{relativepos}
Carl Doersch, Abhinav Gupta, and Alexei~A Efros.
\newblock Unsupervised visual representation learning by context prediction.
\newblock In {\em Int. Conf. Comput. Vis.}, 2015.

\bibitem{doersch2017multi}
Carl Doersch and Andrew Zisserman.
\newblock Multi-task self-supervised visual learning.
\newblock In {\em Int. Conf. Comput. Vis.}, 2017.

\bibitem{carl}
Carl Doersch and Andrew Zisserman.
\newblock Multi-task self-supervised visual learning.
\newblock In {\em Int. Conf. Comput. Vis.}, 2017.

\bibitem{bigan}
Jeff Donahue, Philipp Kr{\"a}henb{\"u}hl, and Trevor Darrell.
\newblock Adversarial feature learning.
\newblock In {\em Int. Conf. Learn. Represent.}, 2017.

\bibitem{bigbigan}
Jeff Donahue and Karen Simonyan.
\newblock Large scale adversarial representation learning.
\newblock In {\em Adv. Neural Inform. Process. Syst.}, 2019.

\bibitem{exemplar}
Alexey Dosovitskiy, Jost~Tobias Springenberg, Martin Riedmiller, and Thomas
  Brox.
\newblock Discriminative unsupervised feature learning with convolutional
  neural networks.
\newblock In {\em Adv. Neural Inform. Process. Syst.}, 2014.

\bibitem{ali}
Vincent Dumoulin, Ishmael Belghazi, Ben Poole, Olivier Mastropietro, Alex Lamb,
  Martin Arjovsky, and Aaron Courville.
\newblock Adversarially learned inference.
\newblock In {\em Int. Conf. Learn. Represent.}, 2017.

\bibitem{gidaris2018unsupervised}
Spyros Gidaris, Praveer Singh, and Nikos Komodakis.
\newblock Unsupervised representation learning by predicting image rotations.
\newblock In {\em Int. Conf. Learn. Represent.}, 2018.

\bibitem{gan}
Ian Goodfellow, Jean Pouget-Abadie, Mehdi Mirza, Bing Xu, David Warde-Farley,
  Sherjil Ozair, Aaron Courville, and Yoshua Bengio.
\newblock Generative adversarial nets.
\newblock In {\em Adv. Neural Inform. Process. Syst.}, 2014.

\bibitem{mganprior}
Jinjin Gu, Yujun Shen, and Bolei Zhou.
\newblock Image processing using multi-code gan prior.
\newblock In {\em IEEE Conf. Comput. Vis. Pattern Recog.}, 2020.

\bibitem{moco}
Kaiming He, Haoqi Fan, Yuxin Wu, Saining Xie, and Ross Girshick.
\newblock Momentum contrast for unsupervised visual representation learning.
\newblock In {\em IEEE Conf. Comput. Vis. Pattern Recog.}, 2020.

\bibitem{resnet}
Kaiming He, Xiangyu Zhang, Shaoqing Ren, and Jian Sun.
\newblock Deep residual learning for image recognition.
\newblock In {\em IEEE Conf. Comput. Vis. Pattern Recog.}, 2016.

\bibitem{cpc2}
Olivier~J H{\'e}naff, Aravind Srinivas, Jeffrey De~Fauw, Ali Razavi, Carl
  Doersch, SM Eslami, and Aaron van~den Oord.
\newblock Data-efficient image recognition with contrastive predictive coding.
\newblock {\em arXiv preprint arXiv:1905.09272}, 2019.

\bibitem{fid}
Martin Heusel, Hubert Ramsauer, Thomas Unterthiner, Bernhard Nessler, and Sepp
  Hochreiter.
\newblock Gans trained by a two time-scale update rule converge to a local nash
  equilibrium.
\newblock In {\em Adv. Neural Inform. Process. Syst.}, 2017.

\bibitem{autoencoder}
Geoffrey~E Hinton and Ruslan~R Salakhutdinov.
\newblock Reducing the dimensionality of data with neural networks.
\newblock {\em Science}, 2006.

\bibitem{hjelm2019learning}
R~Devon Hjelm, Alex Fedorov, Samuel Lavoie-Marchildon, Karan Grewal, Phil
  Bachman, Adam Trischler, and Yoshua Bengio.
\newblock Learning deep representations by mutual information estimation and
  maximization.
\newblock In {\em Int. Conf. Learn. Represent.}, 2019.

\bibitem{lfw}
Gary~B Huang, Manu Ramesh, Tamara Berg, and Erik Learned-Miller.
\newblock Labeled faces in the wild: A database for studying face recognition
  in unconstrained environments.
\newblock Technical report, Technical Report 07-49, University of
  Massachusetts, Amherst, 2007.

\bibitem{adain}
Xun Huang and Serge Belongie.
\newblock Arbitrary style transfer in real-time with adaptive instance
  normalization.
\newblock In {\em Int. Conf. Comput. Vis.}, 2017.

\bibitem{gansteerability}
Ali Jahanian, Lucy Chai, and Phillip Isola.
\newblock On the "steerability" of generative adversarial networks.
\newblock In {\em Int. Conf. Learn. Represent.}, 2020.

\bibitem{jakab2018unsupervised}
Tomas Jakab, Ankush Gupta, Hakan Bilen, and Andrea Vedaldi.
\newblock Unsupervised learning of object landmarks through conditional image
  generation.
\newblock In {\em Adv. Neural Inform. Process. Syst.}, 2018.

\bibitem{johnson2016perceptual}
Justin Johnson, Alexandre Alahi, and Li Fei-Fei.
\newblock Perceptual losses for real-time style transfer and super-resolution.
\newblock In {\em Eur. Conf. Comput. Vis.}, 2016.

\bibitem{pggan}
Tero Karras, Timo Aila, Samuli Laine, and Jaakko Lehtinen.
\newblock Progressive growing of gans for improved quality, stability, and
  variation.
\newblock In {\em Int. Conf. Learn. Represent.}, 2018.

\bibitem{stylegan}
Tero Karras, Samuli Laine, and Timo Aila.
\newblock A style-based generator architecture for generative adversarial
  networks.
\newblock In {\em IEEE Conf. Comput. Vis. Pattern Recog.}, 2019.

\bibitem{stylegan_github}
Tero Karras, Samuli Laine, and Timo Aila.
\newblock Stylegan - official tensorflow implementation.
\newblock \url{https://github.com/NVlabs/stylegan}, 2019.

\bibitem{adam}
Diederik~P Kingma and Jimmy Ba.
\newblock Adam: A method for stochastic optimization.
\newblock In {\em Int. Conf. Learn. Represent.}, 2015.

\bibitem{vae}
Diederik~P Kingma and Max Welling.
\newblock Auto-encoding variational bayes.
\newblock In {\em Int. Conf. Learn. Represent.}, 2014.

\bibitem{alexnet}
Alex Krizhevsky, Ilya Sutskever, and Geoffrey~E Hinton.
\newblock Imagenet classification with deep convolutional neural networks.
\newblock In {\em Adv. Neural Inform. Process. Syst.}, 2012.

\bibitem{mnist}
Yann LeCun, L{\'e}on Bottou, Yoshua Bengio, and Patrick Haffner.
\newblock Gradient-based learning applied to document recognition.
\newblock {\em Proceedings of the IEEE}, 1998.

\bibitem{fpn}
Tsung-Yi Lin, Piotr Doll{\'a}r, Ross Girshick, Kaiming He, Bharath Hariharan,
  and Serge Belongie.
\newblock Feature pyramid networks for object detection.
\newblock In {\em IEEE Conf. Comput. Vis. Pattern Recog.}, 2017.

\bibitem{pytorch-cifar10}
Kuang Liu.
\newblock Pyotrch cifar10.
\newblock \url{https://github.com/kuangliu/pytorch-cifar.git}, 2019.

\bibitem{parnet}
Shu Liu, Lu Qi, Haifang Qin, Jianping Shi, and Jiaya Jia.
\newblock Path aggregation network for instance segmentation.
\newblock In {\em IEEE Conf. Comput. Vis. Pattern Recog.}, 2018.

\bibitem{celeba}
Ziwei Liu, Ping Luo, Xiaogang Wang, and Xiaoou Tang.
\newblock Deep learning face attributes in the wild.
\newblock In {\em Int. Conf. Comput. Vis.}, 2015.

\bibitem{sift}
David~G Lowe.
\newblock Distinctive image features from scale-invariant keypoints.
\newblock {\em Int. J. Comput. Vis.}, 2004.

\bibitem{matthew2014visualizing}
DZ Matthew and R Fergus.
\newblock Visualizing and understanding convolutional neural networks.
\newblock In {\em IEEE Conf. Comput. Vis. Pattern Recog.}, 2014.

\bibitem{cpc}
Aaron van~den Oord, Yazhe Li, and Oriol Vinyals.
\newblock Representation learning with contrastive predictive coding.
\newblock {\em arXiv preprint arXiv:1807.03748}, 2018.

\bibitem{motionseg}
Deepak Pathak, Ross Girshick, Piotr Doll{\'a}r, Trevor Darrell, and Bharath
  Hariharan.
\newblock Learning features by watching objects move.
\newblock In {\em IEEE Conf. Comput. Vis. Pattern Recog.}, 2017.

\bibitem{alae}
Stanislav Pidhorskyi, Donald Adjeroh, and Gianfranco Doretto.
\newblock Adversarial latent autoencoders.
\newblock In {\em IEEE Conf. Comput. Vis. Pattern Recog.}, 2020.

\bibitem{dcgan}
Alec Radford, Luke Metz, and Soumith Chintala.
\newblock Unsupervised representation learning with deep convolutional
  generative adversarial networks.
\newblock In {\em Int. Conf. Learn. Represent.}, 2016.

\bibitem{sharif2014cnn}
Ali Sharif~Razavian, Hossein Azizpour, Josephine Sullivan, and Stefan Carlsson.
\newblock Cnn features off-the-shelf: an astounding baseline for recognition.
\newblock In {\em IEEE Conf. Comput. Vis. Pattern Recog. Worksh.}, 2014.

\bibitem{interfacegan}
Yujun Shen, Ceyuan Yang, Xiaoou Tang, and Bolei Zhou.
\newblock Interfacegan: Interpreting the disentangled face representation
  learned by gans.
\newblock {\em IEEE Trans. Pattern Anal. Mach. Intell.}, 2020.

\bibitem{shocher2020semantic}
Assaf Shocher, Yossi Gandelsman, Inbar Mosseri, Michal Yarom, Michal Irani,
  William~T Freeman, and Tali Dekel.
\newblock Semantic pyramid for image generation.
\newblock In {\em IEEE Conf. Comput. Vis. Pattern Recog.}, 2020.

\bibitem{vgg}
Karen Simonyan and Andrew Zisserman.
\newblock Very deep convolutional networks for large-scale image recognition.
\newblock In {\em Int. Conf. Learn. Represent.}, 2015.

\bibitem{googlenet}
Christian Szegedy, Wei Liu, Yangqing Jia, Pierre Sermanet, Scott Reed, Dragomir
  Anguelov, Dumitru Erhan, Vincent Vanhoucke, and Andrew Rabinovich.
\newblock Going deeper with convolutions.
\newblock In {\em IEEE Conf. Comput. Vis. Pattern Recog.}, 2015.

\bibitem{cmc}
Yonglong Tian, Dilip Krishnan, and Phillip Isola.
\newblock Contrastive multiview coding.
\newblock {\em arXiv preprint arXiv:1906.05849}, 2019.

\bibitem{ssim}
Zhou Wang, Alan~C Bovik, Hamid~R Sheikh, and Eero~P Simoncelli.
\newblock Image quality assessment: from error visibility to structural
  similarity.
\newblock {\em IEEE Trans. Image Process.}, 2004.

\bibitem{wu2018unsupervised}
Zhirong Wu, Yuanjun Xiong, Stella~X Yu, and Dahua Lin.
\newblock Unsupervised feature learning via non-parametric instance
  discrimination.
\newblock In {\em IEEE Conf. Comput. Vis. Pattern Recog.}, 2018.

\bibitem{instdisc}
Zhirong Wu, Yuanjun Xiong, Stella~X Yu, and Dahua Lin.
\newblock Unsupervised feature learning via non-parametric instance
  discrimination.
\newblock In {\em IEEE Conf. Comput. Vis. Pattern Recog.}, 2018.

\bibitem{xu2020unsupervised}
Yinghao Xu, Ceyuan Yang, Ziwei Liu, Bo Dai, and Bolei Zhou.
\newblock Unsupervised landmark learning from unpaired data.
\newblock {\em arXiv preprint arXiv:2007.01053}, 2020.

\bibitem{higan}
Ceyuan Yang, Yujun Shen, and Bolei Zhou.
\newblock Semantic hierarchy emerges in deep generative representations for
  scene synthesis.
\newblock {\em Int. J. Comput. Vis.}, 2020.

\bibitem{tpn}
Ceyuan Yang, Yinghao Xu, Jianping Shi, Bo Dai, and Bolei Zhou.
\newblock Temporal pyramid network for action recognition.
\newblock In {\em IEEE Conf. Comput. Vis. Pattern Recog.}, 2020.

\bibitem{yosinski2014transferable}
Jason Yosinski, Jeff Clune, Yoshua Bengio, and Hod Lipson.
\newblock How transferable are features in deep neural networks?
\newblock In {\em Adv. Neural Inform. Process. Syst.}, 2014.

\bibitem{lsun}
Fisher Yu, Ari Seff, Yinda Zhang, Shuran Song, Thomas Funkhouser, and Jianxiong
  Xiao.
\newblock Lsun: Construction of a large-scale image dataset using deep learning
  with humans in the loop.
\newblock {\em arXiv preprint arXiv:1506.03365}, 2015.

\bibitem{mtcnn}
Kaipeng Zhang, Zhanpeng Zhang, Zhifeng Li, and Yu Qiao.
\newblock Joint face detection and alignment using multitask cascaded
  convolutional networks.
\newblock {\em IEEE Signal Processing Letters}, 2016.

\bibitem{colorful}
Richard Zhang, Phillip Isola, and Alexei~A Efros.
\newblock Colorful image colorization.
\newblock In {\em Eur. Conf. Comput. Vis.}, 2016.

\bibitem{zhang2017split}
Richard Zhang, Phillip Isola, and Alexei~A Efros.
\newblock Split-brain autoencoders: Unsupervised learning by cross-channel
  prediction.
\newblock In {\em IEEE Conf. Comput. Vis. Pattern Recog.}, 2017.

\bibitem{layoutlearinng}
Weidong Zhang, Wei Zhang, and Jason Gu.
\newblock Edge-semantic learning strategy for layout estimation in indoor
  environment.
\newblock {\em Transactions On Cybernetics}, 2019.

\bibitem{tcdcn}
Zhanpeng Zhang, Ping Luo, Chen~Change Loy, and Xiaoou Tang.
\newblock Facial landmark detection by deep multi-task learning.
\newblock In {\em Eur. Conf. Comput. Vis.}, 2014.

\bibitem{zhao2020makes}
Nanxuan Zhao, Zhirong Wu, Rynson~WH Lau, and Stephen Lin.
\newblock What makes instance discrimination good for transfer learning?
\newblock {\em arXiv preprint arXiv:2006.06606}, 2020.

\bibitem{zhou2015object}
Bolei Zhou, Aditya Khosla, Agata Lapedriza, Aude Oliva, and Antonio Torralba.
\newblock Object detectors emerge in deep scene cnns.
\newblock In {\em Int. Conf. Learn. Represent.}, 2015.

\bibitem{zhou2017places}
Bolei Zhou, Agata Lapedriza, Aditya Khosla, Aude Oliva, and Antonio Torralba.
\newblock Places: A 10 million image database for scene recognition.
\newblock {\em IEEE Trans. Pattern Anal. Mach. Intell.}, 2017.

\bibitem{idinvert}
Jiapeng Zhu, Yujun Shen, Deli Zhao, and Bolei Zhou.
\newblock In-domain gan inversion for real image editing.
\newblock In {\em Eur. Conf. Comput. Vis.}, 2020.

\bibitem{zhuang2019local}
Chengxu Zhuang, Alex~Lin Zhai, and Daniel Yamins.
\newblock Local aggregation for unsupervised learning of visual embeddings.
\newblock In {\em Int. Conf. Comput. Vis.}, 2019.

\bibitem{zou2018layoutnet}
Chuhang Zou, Alex Colburn, Qi Shan, and Derek Hoiem.
\newblock Layoutnet: Reconstructing the 3d room layout from a single rgb image.
\newblock In {\em IEEE Conf. Comput. Vis. Pattern Recog.}, 2018.

\end{thebibliography}
}

\end{document}